\documentclass[conference]{IEEEtran}
\IEEEoverridecommandlockouts
\IEEEoverridecommandlockouts
\usepackage{cite}
\usepackage{amsmath,amssymb,amsfonts}
\usepackage{algorithmic}
\usepackage{graphicx}
\usepackage{textcomp}
\usepackage{xcolor}
\usepackage{bm}
\usepackage{multicol}
\usepackage{multirow}
\usepackage{makecell}
\usepackage{algorithm}

\usepackage{array}

\newcolumntype{C}{>{\centering\arraybackslash}p{1em}}
\usepackage{booktabs}
\usepackage{makecell}
\usepackage{graphicx}

\def\BibTeX{{\rm B\kern-.05em{\sc i\kern-.025em b}\kern-.08em
    T\kern-.1667em\lower.7ex\hbox{E}\kern-.125emX}}
\usepackage{epstopdf}
\AppendGraphicsExtensions{.tif}

\begin{document}

\title{An Adaptive Tensor-Train Decomposition Approach for Efficient Deep Neural Network Compression \\

}

\author{\IEEEauthorblockN{Shiyi Luo}
\IEEEauthorblockA{Computational Science Research Center \\
\textit{San Diego State University}\\
San Diego, USA \\
sluo7066@sdsu.edu}
\and

\IEEEauthorblockN{Mingshuo Liu}
\IEEEauthorblockA{Computational Science Research Center \\
\textit{University of California Irvine}\\
Irvine, USA \\
mingshl5@uci.edu}
\and
\IEEEauthorblockN{Pu Sun}
\IEEEauthorblockA{\textit{Department of Electrical and Computer Engineering} \\
\textit{ University of California, Davis }\\
Davis, USA \\
}

\and
\IEEEauthorblockN{Yifeng Yu}
\IEEEauthorblockA{Department of Mathematics \\
\textit{University of California Irvine}\\
Irvine, USA \\
}
\and
\IEEEauthorblockN{Shangping Ren}
\IEEEauthorblockA{Department of Computer Science \\
\textit{San Diego State University, USA }\\
}
\and
\IEEEauthorblockN{Yu Bai}
\IEEEauthorblockA{Department of Electrical and Computer Engineering \\
\textit{California State University Fullerton}\\
Fullerton, USA }
}

\maketitle

\begin{abstract}

In the field of model compression, choosing an appropriate rank for tensor decomposition is pivotal for balancing model compression rate and efficiency. However, this selection, whether done manually or through optimization-based automatic methods, often increases computational complexity. Manual rank selection lacks efficiency and scalability, often requiring extensive trial-and-error, while optimization-based automatic methods significantly increase the computational burden. To address this, we introduce a novel, automatic, and budget-aware rank selection method for efficient model compression, which employs Layer-Wise Imprinting Quantitation (LWIQ). LWIQ quantifies each layer's significance within a neural network by integrating a proxy classifier. This classifier assesses the layer's impact on overall model performance, allowing for a more informed adjustment of tensor rank. Furthermore, our approach includes a scaling factor to cater to varying computational budget constraints. This budget awareness eliminates the need for repetitive rank recalculations for different budget scenarios. Experimental results on the CIFAR-10 dataset show that our LWIQ improved by 63.2$\%$ in rank search efficiency, and the accuracy only dropped by 0.86$\%$ with 3.2x less model size on the ResNet-56 model as compared to the state-of-the-art proxy-based automatic tensor rank selection method. 



\end{abstract}


\section{Introduction}
Deep learning is very popular in many applications, such as natural language processing (NLP), speech recognition, self-driving cars, and healthcare. In these fields, image classification tasks have achieved excellent results using deep learning technology. Especially when deploying deep learning models on mobile devices, accuracy, compact size, and minimal processing latency are essential requirements.

In recent decades, there has been a trend towards increasingly deeper CNN (Convolutional Neural Network) architectures. Accuracy and feature representation are frequently enhanced by these deeper networks. However, the trade-off is that they always necessitate substantial storage and computational resources to ensure optimal performance. This becomes a significant challenge when deploying CNNs on resource-limited devices. To address this, researchers have explored various optimization techniques, including model compression methods, to reduce the model's computational requirements without significantly sacrificing its performance. Among these, pruning\cite{Xiaotian2022,Xiaotian2023} and quantization~\cite{rastegari2016xnor,LIU2023100680}, have been extensively researched and advanced to minimize model redundancy with varying levels of granularity effectively. Moreover, low-rank compression provides a distinct approach to exploring the intrinsic low-rank structure of CNN models at the structural level, focusing on reducing the size of large-weight matrices or tensors into smaller cores. This approach leads to substantial storage savings and computational efficiency enhanced. Various decomposition-based compression approaches have been proposed to date. These approaches can be classified into two categories: 2-D matrix decomposition-based methods~\cite{klema1980singular} and high-order tensor decomposition-based methods~\cite{tucker1966some,liu2022autonomous,Liu2021ASAP}. The 2-D matrix methods focus on simplifying matrix operations, while high-order tensor methods explore multi-dimensional data representations for more complex compression. 


\textbf{Limitation of Current Model Compression.} The current low-rank decomposition approaches to compress deep neural networks (DNNs) exhibit several limitations considering the practical application:

\underline{First}, previous studies have selected ranks through a series of manual adjustments, which often results in considerable inefficiency, is time-consuming, relies heavily on human expertise, and does not always produce optimal compression.
  
\underline{Second}, many advanced automatic rank determination methods are based on heuristic rank search strategies. While these policies automate parts of the process, they significantly increase the computational burden. Because of the iterative and complex nature of heuristic methods, the CNN model usually takes significant computational resources and time to converge to a suitable result.

\underline{Third}, existing budget-aware rank determination approaches often require repeated searches and training when the budget constraints change. These methods need a lot of time and computing resources, hindering efficiency and lacking flexibility.

 ~\textbf{Technical Contributions and Preview.} To tackle these obstacles and facilitate the broader implementation of compression techniques, in this paper, we propose a novel Budget-Aware Automatic Model Compression Framework for designing compact Deep Neural Networks (DNN). Recognizing the pivotal role of parameter selection in shaping DNN architecture, our work delves deeper into the concept of automatic parameter selection. This concept is based on the proposed algorithm, which facilitates a layer quantitation method for TT Decomposition. The aim is to find the most suitable low-rank structure for the model using TT Decomposition. Overall, the contributions of this paper are summarized as follows: 

 \begin{itemize}

\item Researchers are currently delving into the critical impact of parameter selection, such as rank selection for TT Decomposition, on the accuracy of model compression. Nevertheless, a clear understanding of the intricate relationship between parameter selection and model compression accuracy is still a challenging endeavor. Addressing this gap, our paper makes the first attempt to methodically assess this issue. We develop the novel Layer-Wise Imprinting Quantitation (LWIQ) method, designed to precisely evaluate the importance of each layer in a TT decomposed model based on its contribution to accuracy. This innovative approach enables a more informed and precise adjustment of tensor rank, considering the specific requirements of the model structure and data characteristics.
     
    \item Utilizing the proposed LWIQ method, we have devised an automatic rank selection algorithm that streamlines the process of assigning ranks to each layer more efficiently. Unlike other methods that rely on heuristic rank search strategies, our algorithm stands out as it automatically assigns ranks to each layer without requiring intensive searching/trial efforts. Moreover, incorporating a scaling factor adapts to different budgetary constraints without requiring additional rank searches. 
  
    \item We have developed a time-efficient variant of our method, known as LWIQ-sample, which capitalizes on the 'low-shot' nature of our weight imprinting technique. This approach means that our method requires minimal data to accurately assess the importance of each layer. LWIQ-sample facilitates rank search using only 50$\%$ of the dataset, significantly reducing rank search time. This reduction in search time is especially crucial for large datasets, where conventional rank search methods are extremely time-consuming. Our  LWIQ-sample method maintains accuracy while offering substantial improvements in efficiency, demonstrating its potential as a practical solution for handling extensive data.

 \end{itemize}

The remainder of this paper is organized as follows: In Section~\ref{sec:RW}, we introduce the related background in model compression, tensor train decomposition, and tensor selections. In Section~\ref{sec:propose},
we demonstrate a Budget-Aware Automatic Framework for Efficient Model Compression using LWIQ.
Section~\ref{sec:ExpRes} presents the experimental results of the proposed method. Finally, Section~\ref{sec:con} concludes the manuscript.

\section{Related Work}
\label{sec:RW}
\subsection{Model Compression}
Model compression plays a crucial role in Deep Learning that seeks to increase the effectiveness of deep neural networks, especially in these resource-constrained devices, like mobile phones and laptops. These devices often have limited processing power and storage capacity, making the efficient execution of complex models a challenging task. Several main compression techniques have been explored to address this challenge, including pruning, quantization, and tensor decomposition.
\textbf{Pruning} is a widely used technique in model compression by setting certain parts of the Convolutional Neural Network's weights to zero. This process can be classified into two categories: unstructured pruning and structured pruning. Unstructured pruning~\cite{han2015deep,zhang2018systematic} is effective in achieving high compression ratios while maintaining accuracy.  However, this method encounters challenges in achieving theoretical speedup when implemented on practical hardware devices. The irregularity of the pruned connections makes it difficult to fully leverage hardware optimizations for efficient execution. Structured pruning, contrastingly, offers more hardware-friendly solutions. Techniques such as filter pruning~\cite{luo2017thinet,he2019filter,lin2020hrank,sui2021chip} and channel pruning~\cite{he2017channel,zhuang2018discrimination,peng2019collaborative} introduce structured sparsity patterns that align more effectively with hardware architectures. However, the very imposition of these structured patterns can sometimes limit the extent of compression ratio and accuracy achievable, as they may remove potentially useful connections.

\begin{figure*}[htp]
    \centering
    \includegraphics[width=0.8\linewidth]{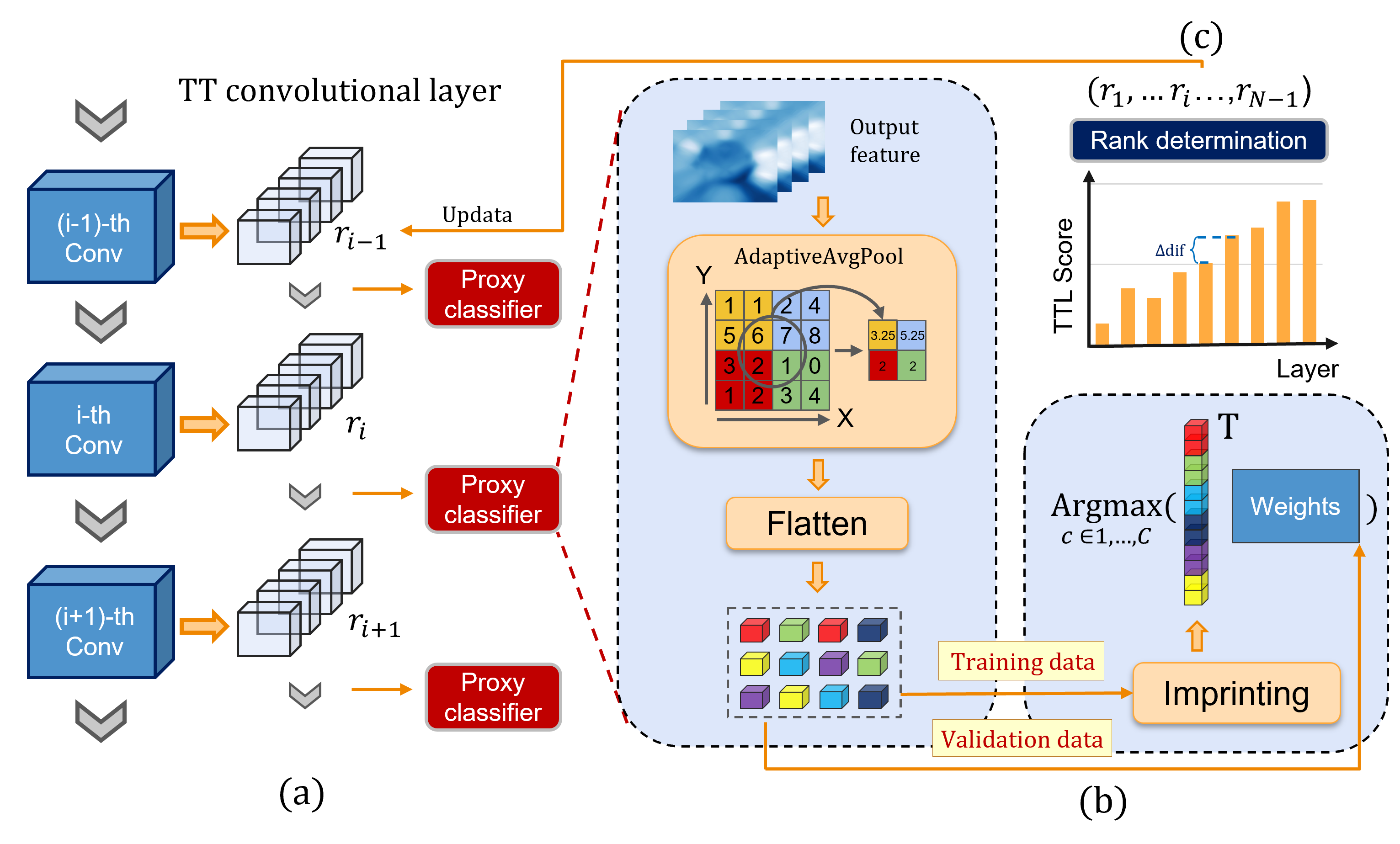}
    \caption{Automatic framework for determining TT's ranks. $(a)$ TT decomposition is applied to every convolutional layer. $(b)$ We insert a proxy classifier to evaluate the layer's unique information. The proxy classifier is estimated by implementing the weight imprinting method. $(c)$ Ranks are determined according to the TT layer importance scores. }
    \label{fig:architecture}
\end{figure*}
\textbf{Model Quantization} is another critical compression approach that involves encoding weights and activations using a limited number of bits. This method has gained widespread adoption for deploying deep neural networks on practical devices~\cite{han2016eie,chen2016eyeriss}. The number precision selected for the quantization scheme is typically determined by balancing the resource budget constraints with the accuracy requirements of the application. In some innovative studies~\cite{rastegari2016xnor,courbariaux2015binaryconnect}, an extreme quantization scheme called 1-bit weight has been proposed. This scheme seeks to minimize storage and computational costs to the utmost by representing weights with just a single bit. While this method offers significant reductions in resource usage, it also poses challenges in maintaining the model's performance, particularly in more complex tasks.

\textbf{Tensor Decomposition}, as a low-rank decomposition, is a critical aspect of model compression. It includes several notable tensor methods, such as the Tucker decomposition~\cite{hoff2016equivariant}, the canonical polyadic decomposition (CPD)~\cite{rai2014scalable,zhao2015bayesian,ermis2014bayesian}, parallel factor (PARAFAC2)~\cite{jorgensen2018probabilistic}, and tensor train decomposition (TT Decomposition)~\cite{liu2021efficient,oseledets2011tensor,deng2019tie}. Each of these methods offers unique advantages in terms of simplifying and reducing the size of large tensors in neural networks. Among them, the TT decomposition stands out due to its advanced properties. The TT decomposition provides an efficient tensor representation in the TT format, significantly reducing memory storage requirements. This efficiency is largely attributed to its utilization of singular-value decomposition, which streamlines the tensor's complexity without sacrificing essential information. Building on these benefits, Garipov et al.~\cite{garipov2016ultimate} proposed an innovative method that applies TT decomposition to both convolutional layers and fully-connected layers of neural networks. This advancement expanded the applicability of TT decomposition, making it an even more versatile tool in the field of model compression.

\subsection{Tensor-Train Decomposition}
Tensor Train decomposition is a powerful method for factorizing a $d$-dimensional tensor, which has dimensions $n_1 \times n_2 \times \cdots \times n_k \times \cdots \times n_d$, into a set of 3-dimensional tensor cores, thereby simplifying complex data structures and reducing storage requirements. Following the naming convention introduced by Garipov et al~\cite{garipov2016ultimate}. Let's consider a tensor $\bm{\mathcal{A}}\in\mathbb{R}^{n_1 \times n_2 \times \cdots \times n_k \times \cdots \times n_d}$. The TT-representation of tensor $\bm{\mathcal{A}}$ is a collection of TT-cores denoted as $\bm{\mathcal{G}}_{:_,n_k,:} \in \mathbb{R}^{r_{k-1}\times n_k \times r_k}$, where $k=1,2,...,d$ and $n_k$ signifies all elements in this dimension. Therefore, the tensor $\bm{\mathcal{A}}$ can be decomposed as follows: $\bm{\mathcal{A}} = \bm{\mathcal{G}}_{:_,n_1,:}\bm{\mathcal{G}}_{:_,n_2,:} \cdots \bm{\mathcal{G}}_{:_,n_d,:}$.
In this decomposition, $r_k$ represents the rank value, which is crucial for simplifying calculations and reducing the decomposition's parameter count. Setting the initial and final rank values, $r_0 = r_d = 1$, allows for the accurate reconstruction of the tensor from its TT-cores. After applying TT decomposition, the number of parameters of tensor $\bm{\mathcal{A}}$ becomes $\sum_{k=1}^d {{r_{k-1}}{n_k}{r_k}}$, which is significantly smaller than the original size $\prod_{k=1}^d n_k$. Thus, to achieve a high compression ratio while maintaining high accuracy, it is important to select a balanced value for $r_k$. By appropriately determining the rank values, TT decomposition enables the representation of the tensor in a compact form while preserving important information.
\begin{figure}[htp]
    \flushleft
    \includegraphics[width=\linewidth]{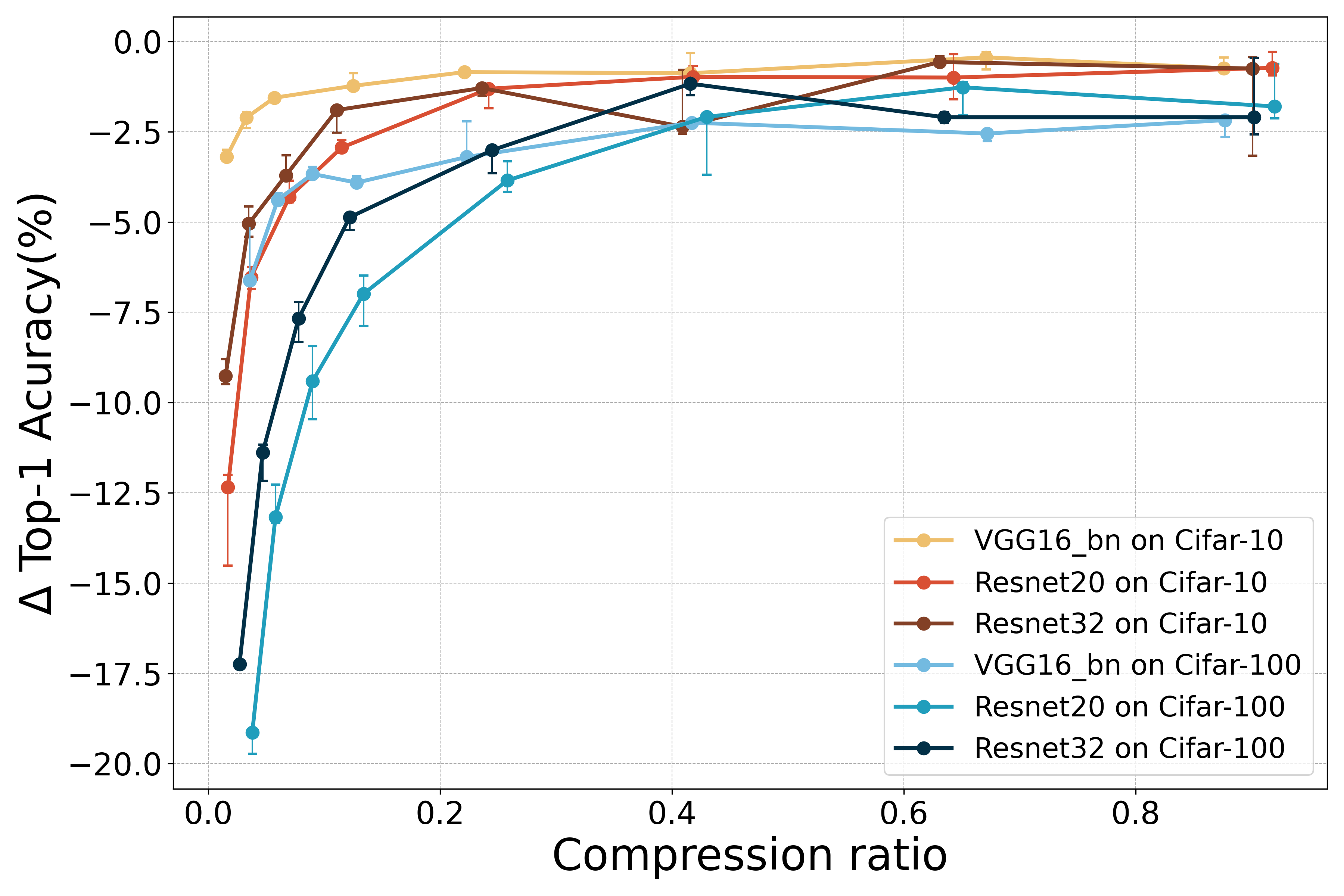}
    \caption{Tensor Train decomposition experiments on various architectures and datasets with baseline comparisons.}
    \label{fig:dif_rank}
\end{figure}

\subsection{Tensor Rank Selection}

The complexity of a model is directly influenced by its rank values, where higher ranks correlate with an increased number of parameters and computational demands. Conversely, the accuracy of the compressed model is also impacted by these rank values. Specifically, low ranks might lead to the loss of crucial information, whereas excessively high ranks could result in overfitting, thereby inflating complexity without proportionate gains in accuracy. Hence, selecting suitable rank values for every layer is a critical and desired task, as shown in the figure~\ref{fig:dif_rank}. To address this challenge, numerous novel techniques have been developed. Optimization-based approaches in tensor factorization and completion often utilize a heuristic called the tensor nuclear norm as a surrogate for tensor rank~\cite{gandy2011tensor,goldfarb2014robust}. Despite their effectiveness, such methods often involve computationally intensive regularization processes on the unfolded tensor, rendering them time-consuming. A promising alternative is to employ Bayesian inference techniques for automatically estimating tensor ranks~\cite{zhao2015bayesian,zhou2013tensor,guhaniyogi2017bayesian}. Bayesian tensor methods offer a powerful solution for automatically estimating tensor ranks based on observed data. These methods not only allow for rank estimation grounded in statistical principles but also accommodate prior knowledge about the data, enhancing their predictive accuracy. However, the application of Bayesian inference to large-scale datasets introduces computational hurdles due to the sophisticated nature of the calculations required. In the realm of TT decomposition for DNN models, a genetic algorithm-based searching strategy has been proposed to iteratively search for the optimal tensor ranks~\cite{li2021heuristic}. Although this strategy offers a systematic approach to finding optimal tensor ranks, its time-consuming nature can limit its applicability, especially in scenarios where efficiency is a critical factor. Researchers continue to explore alternative strategies and optimizations to improve the efficiency of the search process and reduce the overall computational burden. The latest solution, like HALOC~\cite{xiao2023haloc}, is a NAS (neural architecture search) based technique. By leveraging the principles of neural architecture search, they sufficiently explore the rank space and enhance the compatibility of the compression strategy with hardware awareness.

\section{Budget-Aware Automatic Framework for Efficient Model Compression using Layer-Wise Imprinting Quantitation}
\label{sec:propose}
Our research is motivated by the understanding of the relationship between parameter selection and accuracy of the model compression~\cite{elkerdawy2020one,elkerdawy2020filter,liu2021layer,qi2018low}. Thus, it is important to quantify characteristics layer by layer when applying the TT decomposition technique on each convolutional layer. In our approach, we develop a Layer-Wise Imprinting Quantitation (LWIQ) method that provides a quantitation of layer characteristics. This quantification allows our rank selection algorithm to automatically determine an optimal rank for each layer, adjusted by a scaling factor to meet varying computational budgets. In this paper, we demonstrate the effectiveness of the proposed framework for determining the optimal model compression. Our approach involves three steps: 
1)~\textit{Applying TT decomposition.} We initiate our process by decomposing every convolutional layer within the DNN model using Tensor Train decomposition. This step is crucial for reducing the complexity and enhancing the manageability of the model's architecture.
2)~\textit{Embedding layer characteristics calculation module.} Next, we integrate layer characteristics calculation modules into all TT-decomposed layers. These modules leverage proxy classifiers and the weight imprinting method to precisely compute and quantify the characteristics of each layer, providing the data necessary for informed rank selection. Figure~\ref{fig:architecture} illustrates our model architecture, offering insights into the structural implementation of our approach. 
3)~\textit{A budget-aware automatic rank determination algorithm.} Building on the quantified layer characteristics, we propose a novel rank selection strategy. By incorporating a scaling factor, this strategy ensures our model adapts efficiently to various computational budgets, reflecting our commitment to flexibility and optimization. More details are shown on the Algorithm\ref{alg:rank determination}.

\subsection{Weight Imprinting}\label{AA}
The conventional method for estimating the importance of convolutional layers typically involves integrating a classifier head after each layer. This classifier head is composed of adaptive average pooling, followed by a fully connected layer, and finalized with a softmax activation function. For each layer, the weights of these classifiers are trained independently, which, despite its effectiveness, can become a laborious and time-consuming process given the large number of layers in modern deep neural networks. To address this, we adopt the imprinting technique to approximate the weights of the fully connected layer without the need for explicit training. Weight Imprinting initially proposed by Qi et al.\cite{qi2018low}, involves directly setting the final layer weights based on examples from a new category during low-shot learning. We adapt this technique to estimate the weights for the classifier head of each convolutional layer, thereby facilitating a more efficient assessment of layer importance. Compared to training separate classifiers for each layer, imprinting provides a faster and more efficient way. The imprinting approach also is used in other model compression techniques, like pruning\cite{elkerdawy2020one,elkerdawy2020filter} and quantization\cite{liu2021layer}. 

In the described method, we establish a classifier proxy for each convolutional layer to accurately assess the layer's significance. Utilizing the training dataset, we employ the imprinting technique to directly set the weight matrix for each layer's classifier proxy. To standardize the method and achieve a uniform embedding length across layers of varying sizes, adaptive average pooling is applied. This step is crucial for reducing the dimensionality of the output feature map of each layer to a consistent size, facilitating a more straightforward comparison of layer importance. The adaptive average pooling process is mathematically defined as:
\begin{equation}
\begin{aligned}
    &\bm{d}_i = round(\sqrt{\frac{\bm{N}}{n_i}})\\
    &\bm{P}_i =   AdaptiveAvgPool(\bm{O}_i,d_i)
\end{aligned}
\end{equation}
where for any given layer $i$, $n_i$ represents the output feature channels. The embedding length $\bm{N}$ is a predefined value. The function AdaptiveAvgPool is then applied to the output feature map $\bm{O}_i$ of layer $i$. This operation effectively reduces $\bm{O}_i$ to an embedding $\bm{P}_i$.

After the adaptive average pooling operation, the resulting layer embeddings are flattened, preparing them for the imprinting process.  The imprinting technique then calculates the weight matrix $\bm{F}_i$ for each layer $i$'s proxy classifier according to the formula:
\begin{equation}
\begin{aligned}
    \bm{F}_i[:,c]=\frac{1}{\bm{N}_c}\sum_{s=1}^S{\mathbb{I}_{[c_s==c]}\bm{P}_s}
\end{aligned}
\end{equation}
Here, $\bm{F}_i[:,c]$ denotes the weight matrix of the proxy classifier for class $c$ in layer $i$. $\bm{N}_c$ represents the total count of training samples belonging to class $c$. $S$ is the total number of samples in the training data. $\mathbb{I}_{[.]}$ denotes the indicator function, which is 1 if the condition inside the brackets is true and 0 otherwise. $\bm{P}_s$ is the flattened embedding of sample $s$ from layer $i$.  This imprinting process effectively captures the collective influence of samples from each class on the layer's output.

By implementing this imprinting process to each layer of the convolutional neural network, we efficiently determine the weights for the proxy classifiers. These weights are pivotal for assessing the relative importance or ranking of each layer within the network, which is based on their contribution to the overall task performance. The prediction for each sample $s$ is calculated for each layer $i$ using the following equation:
\begin{equation}
    \hat{\bm{y}}_s=arg\max_{c \in {1,...,C}}{ \bm{F}_i[:,c]^T \bm{P}_s}
\end{equation}

\begin{figure}[htp]
    \flushleft
    \includegraphics[width=\linewidth]{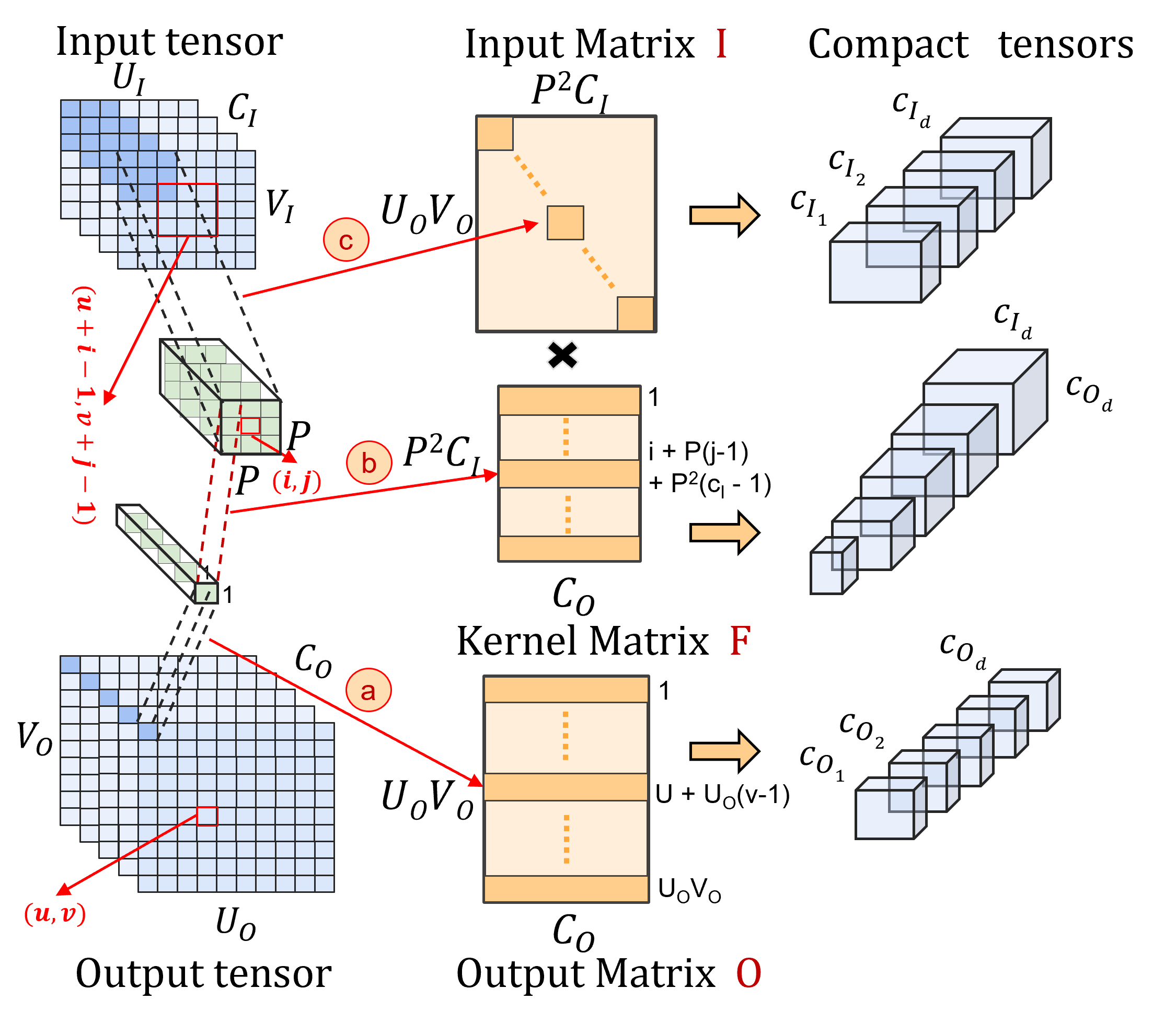}
    \caption{TT convolutional layer. The stages (a), (b), and (c) represent the process that converts the input, output, and kernel tensors into matrices.} 
    \label{fig:TT_layer}
\end{figure}

\subsection{Tensor-Train Decomposed Convolutional Layer}

Garipov et al.\cite{garipov2016ultimate} highlighted a critical limitation in directly applying TT decomposition to factorize the convolutional kernel into a series of low-rank matrices, primarily due to the unique structure and operational dynamics of convolutional layers. To address this challenge, they proposed a novel decomposition technique that effectively adapts to both convolutional and fully connected layers. This innovative decomposition approach provides a solution that can effectively handle the intricacies of convolutional layers, allowing for better compression and optimization of these layers in neural network models. The process can be divided into three main steps:  \emph{Tensor to Matrix Conversion}, \emph{Matrix to Compact Tensor Transformation}, and \emph{TT Decomposition Application}.

\textbf{Convolutional Layer} In a convolutional layer, the input tensor $\bm{\mathfrak{I}}$ with dimensions $ \mathbb{R}^{{\bm{U}}_{I} \times {\bm{V}}_{I} \times {\bm{C}}_{I}}$, and the output tensor $\bm{\mathfrak{O}}$ with dimensions $\mathbb{R}^{{\bm{U}}_{O} \times {\bm{V}}_{O} \times {\bm{C}}_{O}}$ the spatial dimensions (width and height) and the number of channels of the input and output feature maps, respectively. Similarly, the 4-dimensional kernel tensor can be written as $\bm{\mathfrak{F}}\in \mathbb{R}^{{\bm{P}} \times {\bm{P}} \times {\bm{C}}_{I}\times {\bm{C}}_{O}}$. Here, $P \times P$ denotes the size of the convolutional kernel (also known as the filter size), and $\bm{C}_I$ and $\bm{C}_O$ denote the number of input and output channels. Let us define the kernel tensor elements as $\bm{\mathfrak{F}}(i,j,c_I,c_O)$. The operation of a convolutional layer is summarized as follows: 
\begin{equation}
\begin{aligned}
\bm{\mathfrak{O}}(u,v, c_{O})= \sum_{i=1}^P \sum_{j=1}^P \sum_{c_I=1}^{C_I} &\bm{\mathfrak{F}}(i,j,c_I,c_O)\\ &\bm{\mathfrak{I}}(u+i-1,v+j-1,c_I)
\end{aligned}
\end{equation}

\textbf{First-step}: \emph{Reshape tensor convolution to the matrix multiplication}.To facilitate the analysis of the convolution process, we initially convert the input tensor $\bm{\mathfrak{I}}$ of size ${{\bm{U}}_I} \times {\bm{V}}_{I} \times {\bm{C}}_{I}$ and the output tensor $\bm{\mathfrak{O}}$ of size ${\bm{U}}_{O} \times {\bm{V}}_{O} \times {\bm{C}}_{O}$ into their matrix equivalents. We start by defining a patch from the input tensor with dimensions $P \times P \times {\bm{C}}_{I}$ and a corresponding output tensor patch of size $1 \times 1 \times {\bm{C}}_{O}$. From this, we can easily establish the dimensional relationships ${\bm{V}}_{O}={\bm{V}}_{I} - P + 1$ and ${\bm{U}}_{O}={\bm{U}}_{I} - P + 1$. Next, we transform the output tensor $\bm{\mathfrak{O}}$ into a matrix $\bm{O}$ via the following mapping: $\bm{\mathfrak{O}}(u,v,c_O) = \bm{O}(u+{\bm{U}}_{O}(v-1),c_O)$. As depicted in Figure~\ref{fig:TT_layer}(a). Here, $u+{\bm{U}}_{O}(v-1)$ represents the mapping position of the patch $1 \times 1 \times {\bm{C}}_{O}$ in the matrix $\bm{O}$, which corresponds to the height of the output matrix $\bm{O}$. $c_O$ indicates the width of the matrix $\bm{O}$. Similarly, the input tensor $\bm{\mathfrak{I}}$ and kernel tensor $\bm{\mathfrak{F}}$ can be reshaped as follows:
\begin{equation}
\begin{aligned}
  &\bm{\mathfrak{I}}(v+i-1,u+j-1,c_I)= \\ &\bm{I}(u+{\bm{U}}_{O}(v-1),i+P(j-1)+P^2(c_I-1))\\
  &\bm{\mathfrak{F}}(i,j,c_I,c_O)=\bm{F}(i+P(j-1)+P^2(C_I-1),c_O)\\
\end{aligned}
\end{equation}
Here, $i$ and $j$ range from 1 to $P$. Figure~\ref{fig:TT_layer}(b)(c) provides a detailed description of the process that converts the tensor into matrix multiplication, illustrating the corresponding relationships between the input tensor, output tensor, and filter. By transforming the tensors into matrices using the defined mappings, this reshaping operation rearranges the elements of the tensor into a 2-dimensional matrix structure, enabling subsequent processing using matrix operations. 

\begin{figure*}[htp]
    \centering
    \includegraphics[width=\linewidth]{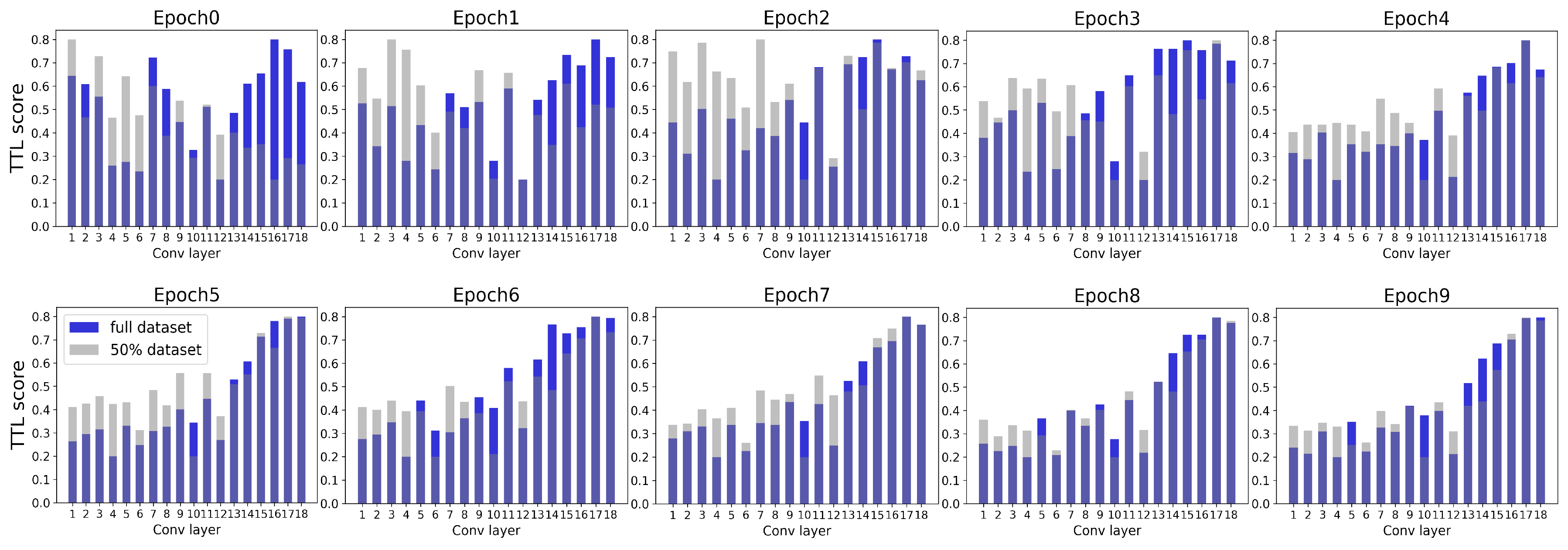}
    \caption{Evolution of layer ranking in iterative imprinting. The model used is ResNet20 on the CIFAR-100 dataset.}
    \label{fig:ttl_score}
\end{figure*}
\textbf{Second-step}: \emph{Transforming the matrix into a more compact tensor by constructing objective functions}. Consider a matrix $\bm{A} \in \mathbb{R}^{M \times N}$, where $M = \prod_{a=1}^d m_a$ and $N = \prod_{a=1}^d n_a$, is reshaped into a more compact tensor $\bm{\mathcal{A}} \in \mathbb{R}^{n_1 m_1 \times n_2 m_2 \times ... \times n_a m_a}$. The reshaping process utilizes two objective functions to map each element of the matrix $\bm{A}$ to the tensor $\bm{\mathcal{A}}$: 
\begin{equation}
\begin{aligned}
&F(i)=[f_1(i),...,f_a(i),...,f_d(i)] \ \ \ \ \ f_a(i) \in (1,...,m_a)\\ 
&G(j)=[g_1(j),...,g_a(j),...,g_d(j)]\ \ \ \  
g_a(j) \in (1,...,n_a)\\
&where\  a \in (1,...,d), i \in (1,...,M), and \ j \in (1,...,N)
\end{aligned}
\end{equation}
Each objective function corresponds to a specific dimension in the tensor, represented by the index $a$. Through this method, the elements $\bm{A}(i,j)$ are represented in the tensor format as $\bm{A}(i,j)=\bm{\mathcal{A}}((f_1(i),g_1(j)),...,(f_a(i),g_a(j)),...,(f_d(i),g_d(j)))$. This notation highlights the relationship between the elements of the matrix $\bm{A}$ and the corresponding elements of the compact tensor $\bm{\mathcal{A}}$. Consequently, the input matrix $\bm{I}$ with size $\bm{U}_O\bm{V}_O \times P^2\bm{C}_I$ and output matrix $\bm{O}$ with size $\bm{U}_O\bm{V}_O \times \bm{C}_O$ are effectively reconstructed into their compact tensor forms:
\begin{equation}
\begin{aligned}
    &\bm{I}(u+{\bm{U}}_{O}(v-1),i+P(j-1)+P^2(c_I-1))=\\
    &\hat{\bm{\mathcal{I}}}(u+i-1,v+j-1,c_{I_1},...,c_{I_a},...,c_{I_d})\\
    &\bm{O}(u+{\bm{U}}_{O}(v-1),c_O)=\\
    &\hat{\bm{\mathfrak{O}}}(u,v, c_{O_1},...,c_{O_a},...,c_{O_d})
\end{aligned}
\end{equation}
where $c_I = \prod_{a=1}^d C_{I_a}$ and $c_O = \prod_{a=1}^d C_{O_a}$, indicating that the input and output channels are decomposed into products of several dimensions, further optimizing the tensor's structure for efficient computation.

\textbf{Third-step}: \emph{Applying TT decomposition to the kernel tensor}. The TT format is utilized for decomposing matrices to further optimize, leveraging the synergies between TT decomposition and low-rank decomposition techniques. Take the matrix $\bm{A}$ as the example, the TT format enables representing $\bm{A}$'s elements in a more structured and compact manner. The decomposition is illustrated as follows:
\begin{equation}
\begin{aligned}
 &\ \ \ \ \ \ \ \ \ \ \ \ \ \ \ \ \ \ \ \ \ \ \ \ \ \ \ \ \ \ \ \ \ \bm{A}(i,j)\\
 &=\bm{\mathcal{A}}((f_1(i),g_1(j)),...,(f_a(i),g_a(j)),...,(f_d(i),g_d(j)))\\
 &={\bm{G}}_1[(f_1(i),g_1(j))]...{\bm{G}}_a[(f_a(i),g_a(j))]... {\bm{G}}_d[(f_d(i),g_d(j))]
  \end{aligned}\label{con:TT-format}
\end{equation}
In this equation, ${\bm{G}}_a[(f_a(i),g_a(j))]$ denotes the sub-tensor associated with the indices $(f_a(i),g_a(j))$ for the $a$-th dimension in the TT decomposition. This approach enables the elements of $\bm{A}$ to be expressed as a product of the corresponding sub-tensors, yielding a compact representation conducive to efficient computation and storage.

To apply TT decomposition to the convolutional kernel tensor, we represent the kernel as follows:
\begin{equation}
\begin{aligned}
  &\bm{F}(i+P(j-1)+P^2(\hat{c_I}-1), \hat{c_O})=\\
  &\hat{\bm{\mathfrak{F}}}((i+P(j-1),1),(c_{I_1},c_{O_1}),...,(c_{I_a},c_{O_a}),...,(c_{I_d},c_{O_d}))=\\
  &\hat{\bm{G}_0}[i+P(j-1),1]\bm{G}_1[c_{I_1},c_{O_1}]...\bm{G}_a[c_{I_a},c_{O_a}]...\bm{G}_d[c_{I_d},c_{O_d}]\\
  &where \ \hat{c_I}=c_{I_1}+\sum_{i=2}^d (c_{I_i}-1)\prod_{j=1}^{i-1} c_{I_j}\\  
  &\ \ \ \ \ \ \ \ \ \hat{c_O}=c_{O_1}+\sum_{i=2}^d  (c_{O_i}-1)\prod_{j=1}^{i-1} c_{O_j}
\end{aligned}
\end{equation}
Here, $\hat{\bm{G}_0}$ symbolizes the tensor core corresponding to the convolutional operation, and $\bm{G}_1$ to $\bm{G}_d$ represent the sequential tensor cores that facilitate this decomposition.

\textbf{TT Convolutional layer} After the kernel tensor is decomposed, the convolutional operation within the TT format can be concisely expressed as follows:
\begin{equation}
\begin{aligned}
 &\hat{\bm{\mathfrak{O}}}(u,v, c_{O_1},...,c_{O_d})=\\
 & \sum_{i=1}^P \sum_{j=1}^P \sum_{c_{I_1},..,c_{I_d}}\hat{\bm{\mathfrak{I}}}(u+i-1,v+j-1,c_{I_1},...,c_{I_d})\\
 &\hat{\bm{G}_0}[i+P(j-1),1]\bm{G}_1[c_{I_1},c_{O_1}]...\bm{G}_d[c_{I_d},c_{O_d}]
\end{aligned}
\end{equation}
The work achieves a compact and efficient representation of the original tensor by utilizing the TT format and applying TT decomposition. This approach leverages the low-rank properties of the matrix and exploits the capabilities of TT decomposition to reduce storage requirements and improve computational efficiency.
\begin{algorithm}
  \renewcommand{\algorithmicrequire}{\textbf{Input:}}
	\renewcommand{\algorithmicensure}{\textbf{Output:}}
	\caption{Automatic rank selection algorithm based on LWIQ} 
	\label{alg3} 
	\begin{algorithmic}
		\REQUIRE Training Dataset $Q$, The number of TT convolutional layers $K$, index $k\in[1, K]$, Initial rank for every layer $R$, Training iterations $E={[e_1,e_2,...,e_n]}$\\
		\ENSURE Apply TT decomposition for all convolutional layers $K$ and set the initial rank $R$ for rank decision stage
		\STATE 1: Model trained for $N$ epochs
                \STATE \ \ \ \ $M_{kn}$ = $\bm{train}(R, Q)$, $n \in[1,N]$ and $k \in [1,K]$
            \STATE 2: Weight Imprinting for each TT Layer $k$ in epoch $e_n$
                \STATE \ \ \ \ $Acc_{kn} = \bm{Imprint}(M_{kn})$ 
            \STATE 3: TT convolutional layer scores, as TTL scores $L$
                \STATE \ \ \ \ $L_{kn} = \bm{normalization}(Acc_{kn})$, $\alpha_{1} \leq L_{kn} \leq \alpha_{2}$
            \STATE 4: Consider a group $G_{kn} = [l_{k(n-4)}, l_{k(n-3)}, ... , l_{kn}]$
            \STATE 5: Calculate the standard deviation:\\
                 \STATE \ \ \ \ $\sigma(G_{kn}) = \bm{sqrt}$ ($\frac{1}{5} \sum_{m=n-4}^{n} (l_{km} - \mu(G_{kn}))^2$)\\
                 \STATE\ \ \ \ \ \ \ \ \ \ \ \ \ \ \ \ \  where $\mu(G_{kn})=\frac{1}{5} \sum_{m=n-4}^{n} l_{km}$
            \STATE 6: A layer $k$ at epoch $e_n$ is considered stable if $\sigma(G_{kn}) < \beta$
                
            \STATE 7: At each epoch $e_n$, count the number of stable layers $M_i$\\
                \ \ \ \ training stops at epoch $N$ if $M_n > 0.8 K$
                
            \STATE 8: Final TTL score decision 
            \IF{$k$ in $0.8 K$}
                \STATE $F_{kn}$ = $\mu(G_{kn})$ if $\sigma({G_{kn})<\beta}$
             \ELSIF{$k$ in remaining $0.2K$} 
                \STATE $F_k$ = $\mu(G_{k})$ for epochs $N-4$ to $N$
             \ENDIF
         
            \STATE 9: Rank determination, scaling factor $\gamma$\\
                \ \ \ \  $\bm{R} = \gamma(F_1, F_2,...,F_k,..., F_K)$
            \STATE 10: Update $R_k$ using $\bm{R}$, and fine-tune.
        \end{algorithmic}
        \label{alg:rank determination} 
\end{algorithm}

For the parameter calculation, we suppose the size of the original kernel tensor is $P \times P \times {\bm{C}}_{I} \times {\bm{C}}_{O}$. After applying the traditional tensor train decomposition approach, the size of the kernel tensor becomes $(r_0 \times P \times r_1) + (r_2 \times P \times r_3) + (r_4 \times \bm{C}_I \times r_5) + (r_6 \times \bm{C}_O \times r_7)$. In our compact convolutional layer, the number of parameters of the kernel tensor can be calculated as $(r_0 \times P^2 \times r_1) + (r_2 \times c_{I_1} \times c_{O_1} \times r_3) + (r_4 \times c_{I_2} \times c_{O_2} \times r_5) + (r_6 \times c_{I_3} \times c_{O_3} \times r_7) + (r_8 \times c_{I_4} \times c_{O_4} \times r_9)$. Let us consider the kernel tensor with size $3 \times 3 \times 64 \times 64$ as an example. The rank values are set to $2$, and the first and last rank values should be $1$. So the original size is $36864$, and the traditional TT method size is $402$.  By applying TT decomposition, the input channel and output channel number $64$ can be presented by $(4 \times 4 \times 2 \times 2)$, so the number of parameters for the compact TT method is $170$, which significantly decreases the kernel tensor size compared with the original value of $36846$.

\subsection{Automatic rank determination}
We propose a novel approach for automatically selecting ranks for TT decomposition. Instead of traditional optimization-based approaches that may take long convergence times to approach the optimal solution, our method introduces a unique strategy based on our proposed LWIQ method. Our strategy begins with assigning a uniform rank value $R$ to all TT convolutional layers during the initial rank decision stage. This initial assignment facilitates a pre-convergence phase, effectively enabling an early assessment of the importance of each TT convolutional layer.  Following normalization, we derive an importance score for each layer, termed the TTL score. It's worth mentioning that our TTL score operates on a relative basis, which means the model doesn't need to achieve full convergence. We could terminate the pre-convergence stage and determine rank values as long as the relative TTL scores tend towards stability, which offers substantial time savings. 

The selection of lower rank values, while beneficial for reducing parameter count, may adversely affect model accuracy. Conversely, opting for higher rank values can bolster accuracy, but at the cost of increased computational demand, as illustrated in figure~\ref{fig:dif_rank}. To address this issue, we propose a straightforward policy: assign relatively higher ranks to layers with a high TTL score to ensure more important information is retained. Conversely, allocate relatively lower ranks to layers with low TTL scores to boost compression efficiency. This policy ensures a balanced approach, retaining vital information in key layers while optimizing overall model efficiency.
\begin{figure}[htp]
    \flushleft
    \includegraphics[width=\linewidth]{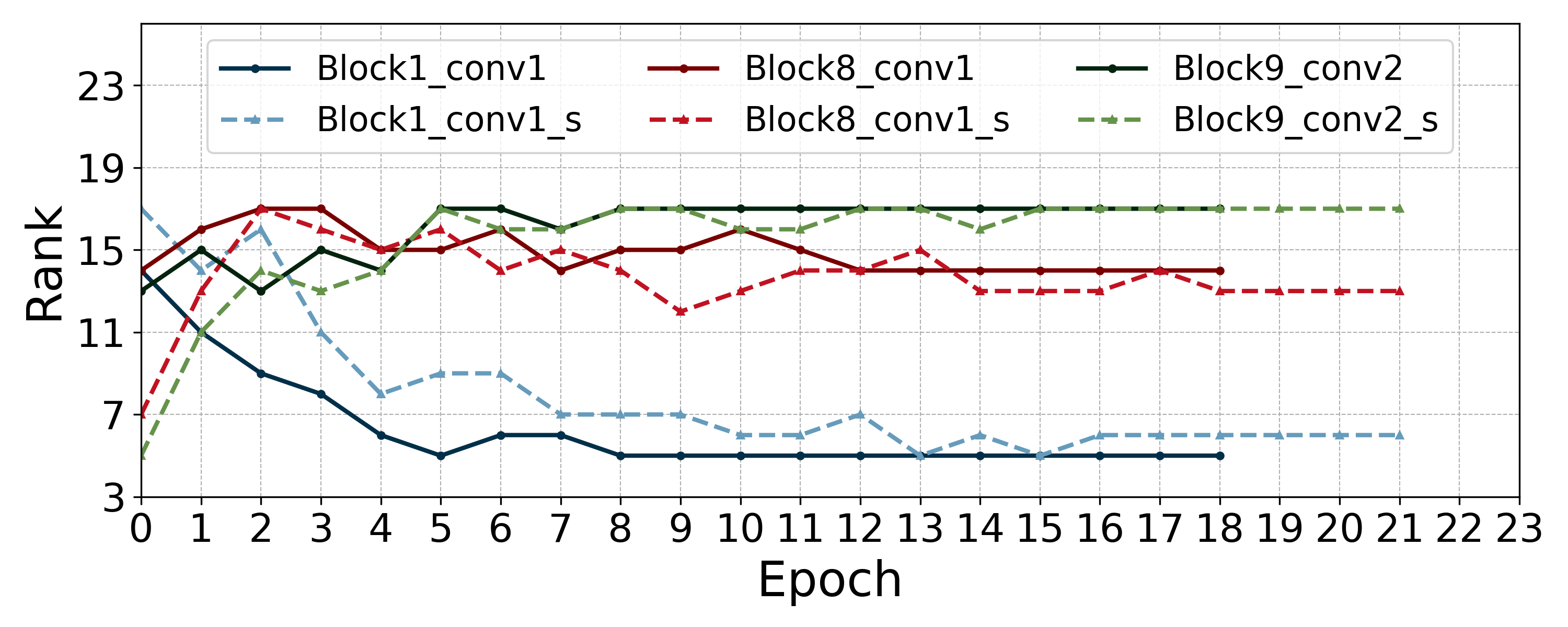}
    \caption{Rank variations for three different convolutional layers of Resnet20 on CIFAR100 and sampled CIFAR100 during training.}
    \label{fig:rank_epoch}
\end{figure}

\begin{figure}[htp]
    \flushleft
    \includegraphics[width=\linewidth]{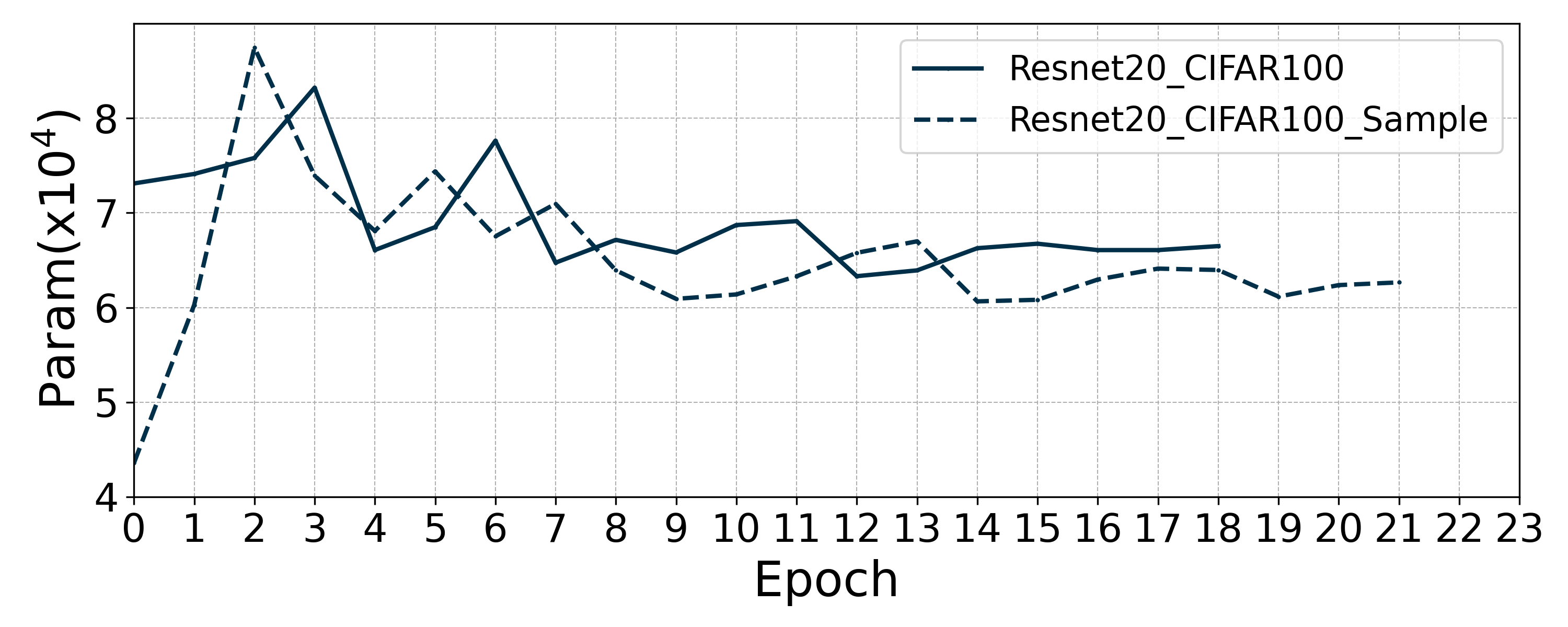}
    \caption{The number of parameters in the rank decision process corresponds to the Fig~\ref{fig:rank_epoch}.}
    \label{fig:param_epoch}
\end{figure}
Algorithm~\ref{alg3} outlines our novel rank determination strategy employed during pre-convergence. Let us define TTL scores as an array $L_k={[l_{k1}, l_{k2},..., l_{kn}]}$ for each epoch $e_i$ and the corresponding layer $k$ in a CNN model. To finalize the TTL scores, we employ three criteria: \textit{Layer Stability Decision}, \textit{Training Termination Decision}, and \textit{Final Value Decision}. For \textit{Layer Stability Decision}, we utilize the standard deviation as a metric to assess the stabilization of TTL scores across training epochs. A hyperparameter $\beta$, serves as the decision threshold aids in this determination. For layer $k$, we analyze the standard deviation of TTL scores grouped into five epochs, $G_{kn} = [l_{k(n-4)}, l_{k(n-3)}, ... , l_{kn}]$, updating with the latest epoch and discarding the oldest to maintain a constant group size. When the standard deviation $\sigma(G_{kn})$ falls below $\beta$. A layer's TTL score is deemed stable when its standard deviation $\sigma(G_{kn})$ drops below $\beta$, with the final TTL score $F_{kn}$ calculated as the group's average. 

\textit{Training Termination Decision} comes into play when $80\%$ of the total layers $K$ reach a standard deviation below $\beta$, indicating sufficient stability to conclude the rank decision phase and avoid unnecessary computation due to the instability of a few layers. For the remaining $20\%$ of the layers, their final TTL scores are derived using the epoch group available at training termination, ensuring all layers are accounted for in the rank determination process. Addressing diverse hardware resource constraints, we introduce a scaling factor ($\gamma$) in our \textit{Final Value Decision}, allowing for adjustments in the model's parameter configuration to achieve various compression ratios.

Our LWIQ, utilizing the weight imprinting method along with our three criteria on TT decomposition, adaptively obtained the final TTL scores and determined the rank values for each convolutional layer with a DNN model. The TTL scores reflect the unique interplay between the model’s architecture and the dataset it is trained on, facilitating the automatic assignment of ranks to layers based on their importance. This approach ensures an optimal balance between the model's accuracy and the number of parameters, enhancing the efficiency of rank selection.

\begin{figure}[htp]
    \flushleft
    \includegraphics[width=\linewidth]{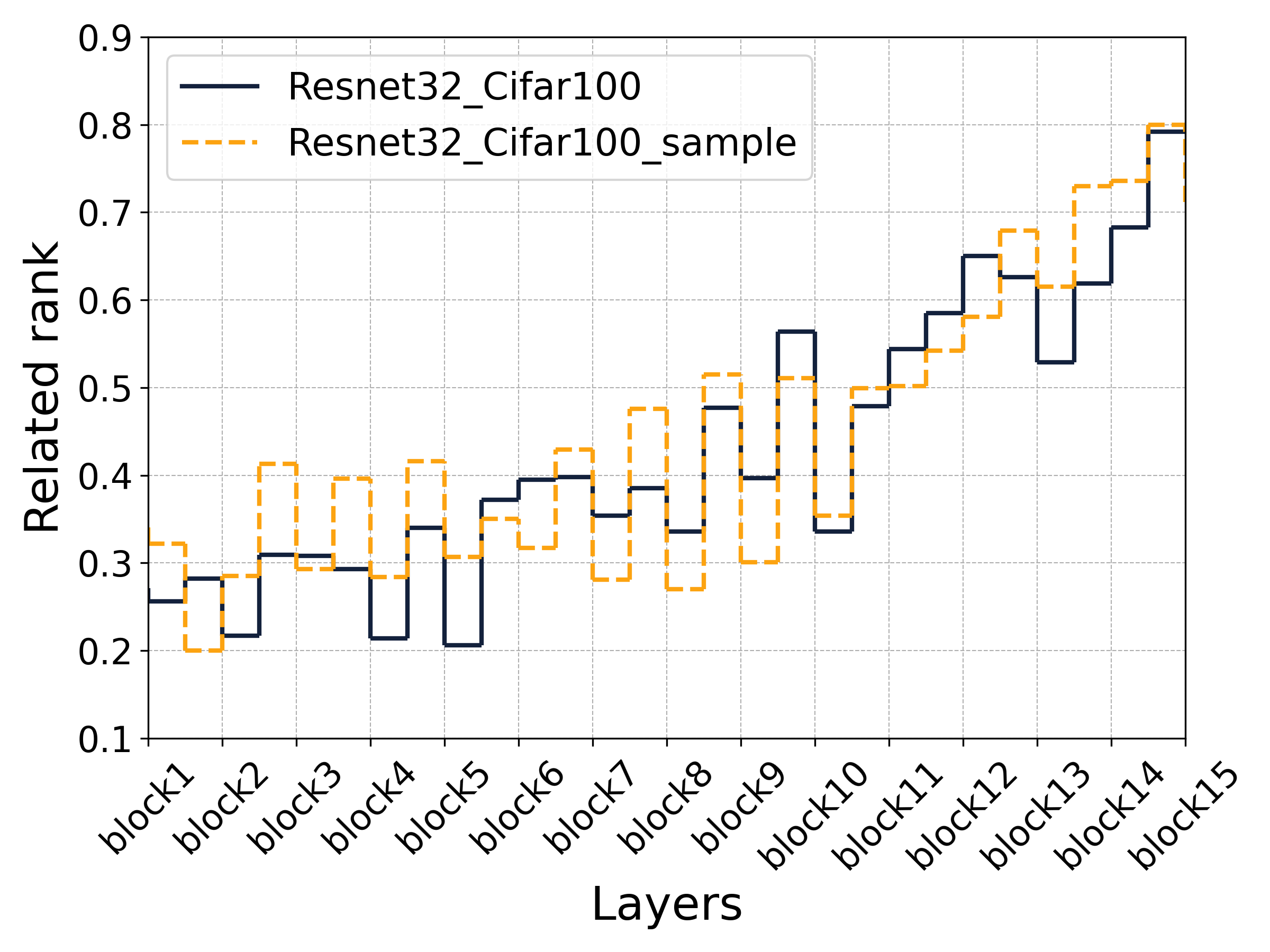}
    \caption{Final rank distribution using LWIQ for compressing Resnet20 on CIFAR10 and sampled CIFAR10 dataset. Every block includes two convolutional layers.}
    \label{fig:layer_rank}
\end{figure}

\subsection{Dataset sampling}
Utilizing the LWIQ technique allows us to identify the relative importance of different layers for informed TT rank allocation. Because of the ’low-shot’ nature of our weight imprinting technique, using a full dataset is not always necessary in the rank decision stage; a partial dataset sample can still effectively examine data characteristics, leading to faster analysis and reduced computational load in resource-limited devices. Therefore, our experiment sampled $50\%$ of the dataset as a more time-saving way to make comparisons.

\begin{table}[h]
    \centering
    \caption{Gamma variation for the ResNet-32 on the CIFAR-100 dataset}
    \label{tbl: Ablation Study}
    \begin{tabular}{|c|c|c|c|}
        \hline
        \multirow{1}{*}{Model} & Param$\downarrow$ & $\gamma$ & Accuracy(Top-1)  \\
        \hline 
        Baseline &- &- &68.10\\
        \hline
        TT(r=17)\cite{garipov2016ultimate} &\multirow{3}{*}{2x}   &-   &66.16 \\  
        LWIQ  &   &34   &67.94  \\ 
        LWIQ-sample  &  &33   &67.61    \\
        \hline
        TT(r=12)\cite{garipov2016ultimate} &\multirow{3}{*}{4x}   &-   &65.67 \\  
        LWIQ  &   &24   &66.97  \\ 
        LWIQ-sample  &  &24   &66.75    \\
        \hline
        TT(r=10)\cite{garipov2016ultimate} &\multirow{3}{*}{6x}   &-   &64.23 \\  
        LWIQ  &   &20   &65.58  \\ 
        LWIQ-sample  &  &19   &65.02    \\
        \hline
    \end{tabular}
\end{table}
   
\section{Experimental setting and Results}
\label{sec:ExpRes}
Our proposed automatic rank determination method for model compression is evaluated on CIFAR10 and CIFAR100 datasets\cite{krizhevsky2009learning}. In addition, we choose the classical convolutional neural networks ResNet20, ResNet32, and ResNet56\cite{he2016deep} to apply our technique.

\subsection{Experimental Setting}
For training our model, we employ the SGD (Stochastic Gradient Descent) optimizer enhanced with Nesterov momentum of 0.9. The CIFAR dataset's initial learning rate is 0.1, adjusted down by a factor of 0.1 every 30 epochs. We start with an initial rank of 32 during the rank decision phase and fine-tune the model to 200 epochs after applying the determined ranks. The model uses a batch size of 256 and incorporates weight decay at a rate of 0.0005 for regularization. The hyper-parameters $\alpha_1$, $\alpha_2$, and $\beta$ are set to 0.2, 0.8, and 0.02, respectively, to optimize performance. Our experiments are conducted on a GPU RTX 3090.

\begin{table}[h]
\centering
\caption{Search time summary for the ResNet-56 on the CIFAR-10 dataset}
\label{tb2: Search time}
\resizebox{0.48\textwidth}{!}{%
\begin{tabular}{|c|cccc|}
\hline
\multirow{2}{*}{\textbf{Operation}} & \textbf{PARS}\cite{sobolev2022pars} & \textbf{PARS} & \multirow{2}{*}{\textbf{LWIQ}} & \textbf{LWIQ} \\
&\textbf{(w/o f-t)} &\textbf{(w f-t)} &  &\textbf{(sample)} \\
\hline 
Search &35min   & 35min  &\multirow{2}{*}{26min}  &\multirow{2}{*}{16min}  \\  
Model decomposition  &28min   &28min  &  & \\ 
BN calibration  &9min  &9min  &-  &-  \\
Accuracy evaluation  &4min  &4min  &2min  &2min  \\
\hline \hline
\textbf{Total search time}  &1h16min  &1h16min  &28min  &18min  \\
\hline
\textbf{Search epoch}  &200  &200  &21  &24  \\
\hline
\textbf{Fine-Tune epoch}  &-  &51  &200  &200  \\
\hline
\textbf{Device}  &A-100  &A-100  &RTX3090  &RTX3090  \\
\hline
\textbf{FLOPs$\downarrow$}  &2x  &2x  &2x  &2x  \\
\hline
\textbf{Params$\downarrow$}  &-  &-  &3.2x  &3.6x  \\
\hline
\textbf{Top-1(\%)}  &91.2  &93.2  &92.34  &91.83  \\
\hline
\end{tabular}%
}
\end{table}


\begin{table*}
    \centering
    \caption{Comparison with different tensor decomposition-based approaches for ResNet-20 and ResNet-32 on CIFAR10.}
    \label{tb3: Cifar10}
    \begin{tabular}{l c c c c c c}
        \hline
        \multicolumn{1}{c}{\textbf{Model}} & \multicolumn{1}{c}{\textbf{Compression}} & \multicolumn{1}{c}{\textbf{Rank Selection}} & \textbf{Rank Decision} & \multicolumn{1}{c}{\textbf{Top-1(\%) \#Param.$\downarrow$}} & \textbf{Top-1(\%) \#Param.$\downarrow$} & \\ 
        &\textbf{Method} &\textbf{Algorithm} & \textbf{Epoch} & \multicolumn{1}{c}{ResNet20} & ResNet32 & \\ 
        \hline
        \hline
        \multirow{1}{*}{} Original&- &- &-  &91.25\ \ \ \ \ \ \ 1.00x &92.49\ \ \ \ \ \ \ 1.00x & \\
        \hline
        \multirow{7}{*}{} Tensor Train\cite{garipov2016ultimate}&TT &Fixed &- &89.48\ \ \ \ \ \ \  2.71x &89.32\ \ \ \ \ \ \  2.45x & \\
        \ ALDS\cite{liebenwein2021compressing}&SVD &- &- &90.92\ \ \ \ \ \ \ 3.97x &- \ \ \ \ \ \ \ \ \ \ \ - & \\
        \ LCNN\cite{idelbayev2020low}&SVD &- &- &90.13\ \ \ \ \ \ \ 2.89x &- \ \ \ \ \ \ \ \ \ \ \ - & \\
        \ PSTRN-S\cite{li2021heuristic}&Tensor\ Ring &Genetic\ Algorithm &- &90.80\ \ \ \ \ \ \ 2.50x &91.44\ \ \ \ \ \ \ 2.70x & \\
        \ BATUDE-S\cite{yin2022batude}&Tucker &BC-ADL &100 &91.02\ \ \ \ \ \ \ 2.60x &92.18\ \ \ \ \ \ \ 2.80x & \\
        \textbf{LWIQ(Ours)}&TT &Imprinting &16-19 &90.84\ \ \ \ \ \ \ 2.88x &91.57\ \ \ \ \ \ \ 2.95x & \\
        \textbf{LWIQ-sample(Ours)}&TT &Imprinting &23-26 &90.34\ \ \ \ \ \ \  2.59x &91.45\ \ \ \ \ \ \ 2.69x & \\
        \hline
        \ PSTRN-M\cite{li2021heuristic}&Tensor\ Ring &Genetic\ Algorithm &- &88.50\ \ \ \ \ \ \ 6.80x &90.60\ \ \ \ \ \ \ 5.80x & \\
        \ BATUDE-M\cite{yin2022batude}&Tucker &BC-ADL &100 &89.47\ \ \ \ \ \ \ 6.80x &91.47\ \ \ \ \ \ \ 5.80x & \\
        \textbf{LWIQ(Ours)}&TT &Imprinting &16-19 &88.88\ \ \ \ \ \ \ 6.81x &90.72\ \ \ \ \ \ \ 5.10x & \\
        \textbf{LWIQ-sample(Ours)}&TT &Imprinting &23-26 &88.38\ \ \ \ \ \ \  7.03x &90.68\ \ \ \ \ \ \ 5.39x & \\
        \hline\\
    \end{tabular}
\end{table*}

\begin{table*}
    \centering
    \caption{Comparison with different tensor decomposition-based approaches for ResNet-20 and ResNet-32 on CIFAR100. }
    \label{tb3: Cifar100}
    \begin{tabular}{l c c c c c c}
        \hline
        \multicolumn{1}{c}{\textbf{Model}} & \multicolumn{1}{c}{\textbf{Compression}} & \multicolumn{1}{c}{\textbf{Rank Selection}} & \textbf{Rank Decision} & \multicolumn{1}{c}{\textbf{Top-1(\%) \#Param.$\downarrow$}} & \textbf{Top-1(\%) \#Param.$\downarrow$} & \\ 
        &\textbf{Method} &\textbf{Algorithm} & \textbf{Epoch} & \multicolumn{1}{c}{ResNet20} & ResNet32 & \\ 
        \hline
        \hline
        \multirow{1}{*}{} Original&- &- &-  &65.40\ \ \ \ \ \ \ 1.00x &68.10\ \ \ \ \ \ \ 1.00x & \\
        \hline
        \multirow{7}{*}{} Tensor Train\cite{garipov2016ultimate}&TT &Fixed &- &62.76\ \ \ \ \ \ \ 2.96x &65.79\ \ \ \ \ \ \ 2.40x & \\
        \ PSTRN-S\cite{li2021heuristic}&Tensor\ Ring &Genetic\ Algorithm &- &66.13\ \ \ \ \ \ \ 2.30x &68.05\ \ \ \ \ \ \ 2.40x & \\
        \ BATUDE-S\cite{yin2022batude}&Tucker &BC-ADL &100 &66.67\ \ \ \ \ \ \ 2.80x &68.95\ \ \ \ \ \ \ 2.60x & \\
        \textbf{LWIQ(Ours)}&TT &Imprinting &15-19 &65.21\ \ \ \ \ \ \ 2.32x &67.15\ \ \ \ \ \ \ 2.91x & \\
        \textbf{LWIQ-sample(Ours)}&TT &Imprinting &17-22&64.93\ \ \ \ \ \ \  2.41x &68.20\ \ \ \ \ \ \ 2.77x & \\
        \hline
        \ PSTRN-M\cite{li2021heuristic}&Tensor\ Ring &Genetic\ Algorithm &- &63.62\ \ \ \ \ \ \ 4.70x &66.77\ \ \ \ \ \ \ 5.20x & \\
        \ BATUDE-M\cite{yin2022batude}&Tucker &BC-ADL &100 &64.91\ \ \ \ \ \ \ 4.70x &66.96\ \ \ \ \ \ \ 5.20x & \\
        \textbf{LWIQ(Ours)}&TT &Imprinting &15-19 &63.58\ \ \ \ \ \ \ 4.44x &65.77\ \ \ \ \ \ \ 5.15x & \\
        \textbf{LWIQ-sample(Ours)}&TT &Imprinting &17-22&63.64\ \ \ \ \ \ \  4.57x &65.57\ \ \ \ \ \ \ 5.12x & \\
        \hline\\
    \end{tabular}
\end{table*}
\subsection{Rank Search Process}
In Figure~\ref{fig:ttl_score}, we present the evolution of TTL scores for each convolutional layer within ResNet-20 trained on the CIFAR-100 dataset across the rank determination stage. This observation demonstrates that the relative importance of each layer in ResNet-20 has a trend toward stabilization by epoch 9. This early stabilization highlights the effectiveness of our LWIQ method when applied to both the full dataset and a $50\%$ sampled subset of CIFAR-100, confirming that our method meaningfully discerns the significance of different layers, tailored to the dataset's characteristics and the model's architectural nuances.

Figure~\ref{fig:rank_epoch} illustrates the variation in rank values for three representative convolutional layers within ResNet-20 throughout the rank determination process. Herein, we use a scaling factor $\gamma$ of 22 on both CIFAR-100 and a $50\%$ subset of CIFAR-100. This figure reveals that the rank determination concludes at epoch 18 for CIFAR-100 and epoch 21 for the reduced CIFAR-100 dataset. Complementarily, Figure~\ref{fig:param_epoch} displays the corresponding changes in the number of parameters of ResNet-20. These two figures highlight how each layer's tensor rank and parameter count dynamically adjust over epochs to stabilize.  This adjustment process, influenced by $\gamma$, guides the parameter count toward a target range. Furthermore, Figure~\ref{fig:layer_rank} shows the distribution of the final related ranks for this fully compressed model. The target number of parameters is achieved through the adjustment of the related rank by $\gamma$. This indicates that our method is budget-friendly and adapts to different budgetary constraints without necessitating additional rank
searches.

Table~\ref{tbl: Ablation Study} showcases a performance comparison at various compression levels between traditional TT decomposition and our LWIQ methods, utilizing the ResNet-32 model trained on the CIFAR-100 dataset. Notably, at a 2x compression rate, LWIQ and its variant, LWIQ-sample, report accuracies of 67.94$\%$ and 67.61$\%$ respectively, marking significant improvements of 1.78$\%$ and 1.45$\%$ over the baseline TT approach. This positive trend continues at more aggressive compression rates; for 4x and 6x reductions, LWIQ outperforms TT by margins of 1.30$\%$ and 1.35$\%$. These findings highlight the superior efficiency of the LWIQ methods in achieving optimal balance between model size reduction and accuracy, demonstrating our method's effectiveness and flexibility across different compression scenarios.

\subsection{Search Time Comparison}
In Table~\ref{tb2: Search time}, we detail the search times for training ResNet-56 on both the full CIFAR-10 dataset and a 50$\%$ sample of CIFAR-10. The results reveal LWIQ-sample has a significant reduction in search time by 35.71$\%$ with only a minimal accuracy decrease of 0.51$\%$ under comparable compression ratios to LWIQ. Notably, our LWIQ method outperforms the state-of-the-art method PARS~\cite{sobolev2022pars} that uses the spatial-SVD decomposition method, showing a 63.2$\%$ improvement in time efficiency. Moreover, our LWIQ-sample method further enhances this efficiency, achieving a 76.32$\%$ reduction in search time. Despite the different model compression approaches employed, our LWIQ's total rank selection duration of 28 minutes remains impressively below the 35-minute search time required by PARS. Thus, our LWIQ substantially shortens search times and effectively preserves model accuracy, showcasing remarkable efficiency even on hardware with lower computational capabilities.
 
\subsection{Comparison with Other Rank Selection Methods}
Our Layer-Wise Imprinting Quantitation (LWIQ) method stands out for its efficiency, particularly in the rank selection process for model compression. Table~\ref{tb3: Cifar10} and table~\ref{tb3: Cifar100} present our evaluation on the CIFAR-10 and CIFAR-100 datasets, respectively. Remarkably, our method requires fewer than 30 epochs to identify the optimal ranks substantially less than the 100 epochs needed by the advanced BATUDE method~\cite{yin2022batude}. Specifically, for CIFAR-10, LWIQ finalizes ranks with 16 epochs on ResNet-20 and 19 epochs on ResNet-32; for CIFAR-100, it takes 15 epochs on ResNet-20 and 19 epochs on ResNet-32. The process of our LWIQ-sample extends slightly, requiring 23-26 epochs for CIFAR-10 and 17-22 for CIFAR-100. Despite its swift rank determination, LWIQ consistently delivers high accuracy. For instance, on CIFAR-10 using a ResNet-20 model at a compression ratio of about 2.7×, LWIQ and its sampled version outperform traditional TT methods by achieving accuracy improvements of 1.36$\%$ and 0.86$\%$, respectively. In comparison to the genetic algorithm-based tensor ring approach (PSTRN)~\cite{li2021heuristic}, our LWIQ method exhibits a notable performance improvement, 0.38x higher on compression ratio and 0.04$\%$ higher in accuracy. Furthermore, even compared with the state-of-the-art method BATUDE~\cite{yin2022batude}, our LWIQ only drops 0.18$\%$ accuracy with significant rank selection efficiency, underscoring its effectiveness in optimizing the balance between model compression efficiency and performance.

\section{Conclusion}
\label{sec:con}
In summary, our paper introduces a novel Budget-Aware Automatic Model Compression Framework, revolutionizing DNN optimization for resource-constrained environments. Central to our contributions is the development of Layer-Wise Imprinting Quantitation (LWIQ), a notivative method that evaluates the relative importance of each layer within a DNN after TT compression. This approach allows for more informed and precise adjustments of tensor ranks, considering both model structure and data characteristics. Remarkably, our LWIQ Sample variant dramatically reduces rank search time, especially critical for large datasets, without sacrificing model accuracy. The experimental results demonstrate the effectiveness of our approach, offering substantial improvements in computational efficiency while ensuring optimal performance of compressed models.

For future work, we aim to expand the scope of our experiments by applying our methodologies to different low-rank decomposition techniques, such as Tucker and Tensor Ring decompositions. Additionally, we plan to test the efficacy of our model compression framework on large-scale datasets, with a particular focus on the ILSVRC-2012 dataset~\cite{russakovsky2015imagenet}. Investigating our framework's performance on such a comprehensive and challenging dataset will provide a more robust evaluation of its applicability and effectiveness in real-world scenarios.

\bibliographystyle{IEEEtran}
\bibliography{Ref}

\end{document}